# MTFL: Multi-Timescale Feature Learning for Weakly-supervised Anomaly Detection in Surveillance Videos

Yiling Zhang*, Erkut Akdag*, †, Egor Bondarev, Peter H.N. de With
VCA Group, Department of Electrical Engineering, Eindhoven University of Technology
P.O. Box 513, Eindhoven 5612AZ, The Netherlands
** equal technical contribution †corresponding author:e.akdag@tue.nl*

## ABSTRACT

Detection of anomaly events is relevant for public safety and requires a combination of fine-grained motion information and contextual events at variable time-scales. To this end, we propose a Multi-Timescale Feature Learning (MTFL) method to enhance the representation of anomaly features. Short, medium, and long temporal tubelets are employed to extract spatio-temporal video features using a Video Swin Transformer. Experimental results demonstrate that MTFL outperforms state-of-the-art methods on the UCF-Crime dataset, achieving an anomaly detection performance 89.78% AUC. Moreover, it performs complementary to SotA with 95.32% AUC on the ShanghaiTech and 84.57% AP on the XD-Violence dataset. Furthermore, we generate an extended dataset of the UCF-Crime for development and evaluation on a wider range of anomalies, namely Video Anomaly Detection Dataset (VADD), involving 2,591 videos in 18 classes with extensive coverage of realistic anomalies. *Code, dataset, and features are accessible via* [https://github.com/erktkdg/MTFL](https://github.com/erktkdg/MTFL)

## 1. INTRODUCTION

Nowadays, the smart city concept is a significant research direction that utilizes data collected by a variety of technologies and sensor types to efficiently optimize the delivery of various services and improve life quality for the public. In this setting, anomaly detection in surveillance videos helps to enhance public safety by early detection of threats and incidents. In the current industrial settings, the control-room operators are watching the feeds from multiple surveillance cameras on a display grid. The advances of surveillance technologies have led to a parallel collection of vast amounts of video data, making manual monitoring increasingly impractical. Therefore, automated anomaly detection is a necessity to remove the limitations of manual inspections, while the broad availability of various technologies makes it feasible.

There are two popular approaches towards anomaly detection in the literature. The first one is the binary classifier method, also referred as 'unsupervised', that learns the dominant data representation using normal data only. This approach is often inadequate due to vaguely defined borders of normality in the latent space. Alternatively, weakly-supervised techniques have gained significant popularity, reducing the annotation work by employing video labels. However, the existing anomaly detection and available datasets still face several issues.

A major challenge in anomaly detection is the variable temporal duration of anomalies. In the real world, anomalies occur rarely and vary in duration, leading to complications in learning the temporal dynamics of the events. For instance, burglary and arson anomalies last longer, while littering and traffic accidents are short-duration events. The divergence between short, medium, and long-lived anomalies needs exploration in flexible learning over time, e.g. using time scales. In particular, a model should learn features from data in multiple temporal scales to detect anomalies, regardless of their durations. Another challenge is that currently available datasets provide a limited number of anomalous surveillance videos due to privacy restrictions, resulting in insufficient data for learning anomalous features. Therefore, there is a data gap for specific crucial anomaly use cases (i.e., littering and traffic accidents) of daily urban life. These incidents involving pedestrians, vehicles, and bicycles play a vital role in ensuring public safety and functionality in smart cities.

To overcome the aforementioned challenges, we propose the Multi-Timescale Feature Learning (MTFL) method for anomaly detection. MTFL extracts features at different temporal scales and fuses motion details from short temporal tubelets with contextual information from long tubelets, enhancing the representation capability of anomaly features. The method employs multi-head cross attention (MCA) [1], multi-head self-attention (MSA) [2], and 1D convolution kernel to capture the correlations between multi-timescale features, global temporal dependencies across the three integrated feature levels, and local temporal dependencies within each tubelet, respectively. These factors are utilized for the scaling and fusion of the features to achieve the final feature representation. To address the lack of anomaly video data and to evaluate MTFL on a wider range of anomalies, we extend the UCF-Crime dataset [3] by incorporating data from the Throwing Action dataset [4] and collected road accident videos, resulting in the Video Anomaly Detection Dataset (VADD).

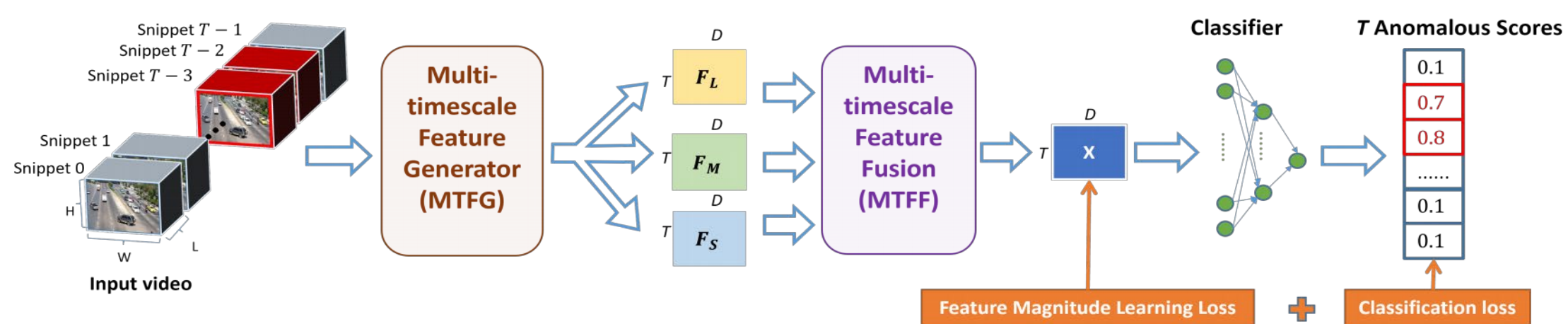

Figure 1. Workflow of Multi-Timescale Feature Learning (MTFL) model. The input video is segmented into $T$ snippets. The Multi-Timescale Feature Generator (MTFG) creates three sets of $T$ features of $D$ dimensions, $\mathbf{F_L}$, $\mathbf{F_M}$, and $\mathbf{F_S}$ , corresponding to features extracted within long, medium, and short temporal tubelets. Next, the Multi-Timescale Feature Fusion (MTFF) captures the correlations among three features and the dependencies among different video snippets to fuse the features into the output feature matrix $\mathbf{X}$. The final anomaly scores of $T$ snippets are obtained after a classifier. A loss function involving feature magnitude loss and classification loss is used for training the MTFF and the classifier.

## 2. RELATED WORK

**Multi-instance Learning Methods**. MIL (multi-instance learning)-based methods under weak supervision consider normal and anomalous videos as 2 bags with short clips of each video as instances in a bag. The separation between abnormal and normal instances is maximized using a ranking loss between the instances with the highest scores in 2 bags. These methods typically utilize a pre-trained network to extract video features, followed by MIL to distinguish these features, such as methods using C3D features [3,5], I3D features [6-11], and Video Swin Transformer features [9,12,13].

**Prediction-based Methods.** Due to the rareness of anomalous videos, the other common approach solves video anomaly detection through one-class prediction or reconstruction that learns normal data only via an encoder-decoder architecture [14-17]. An anomalous segment would lead to a large reconstruction or prediction loss that deviates from the loss of a normal segment. These methods are useful when the anomalous types are complex, unpredictable, and diverse, but the disadvantage is the patterns of normal events are hard to fully define and depend on contextual cues in real life.

**Anomaly Detection Incorporating Attention Mechanisms**. Attention mechanisms [2] determine the temporal or spatial elements to focus on according to the relevance between different data locations, resulting in descriptive and discriminating features well suited for further video anomaly detection. Some recent studies in anomaly detection have applied attention mechanisms and achieved notable results [8,9,14-16,18,19].

**Surveillance Video Datasets for Anomaly Detection**. Although there are some benchmark datasets for anomaly detection, the availability of real-world surveillance anomaly videos and scenarios remains notably constrained. Some datasets contain only normal videos in their training sets and some abnormal events with annotations of bounding boxes in their testing sets, such as UCSD Ped1&2 [17] and ShanghaiTech [20]. Conversely, datasets like UCF-Crime [3], TAD [21], and Throwing Action [4], include various anomalies with video-level labeling in both training and testing sets.

## 3. METHOD

### 3.1 Multi-Timescale Feature Learning (MTFL)

Taking into account the variations in the attention focus of feature information across different frame lengths, the Multi-Timescale Feature Learning (MTFL) method aims to integrate essential features from multiple temporal scales, to enhance the discrimination between abnormal and normal snippets. As illustrated in Fig.1, the input video is divided into $T$ snippets. Each snippet constructs different temporal tubelets based on variable frame lengths, which are used for feature extraction. The Multi-Timescale Feature Generator (MTFG) creates three groups of $T \times D$ features, corresponding to $T$ snippets, namely $\boldsymbol{F_L}$, $\boldsymbol{F_M}$, and $\boldsymbol{F_S}$, long, medium, and short temporal tubelets. Next, the Multi-Timescale Feature Fusion

Network (MTFF) fuses output features obtained from three temporal paths, to create a global feature vector $\boldsymbol{X}$ based on temporal correlations between the snippets. The global vector is generated by multi-head cross and self-attention blocks. Subsequently, we apply a classifier to vector $\boldsymbol{X}$ and generate corresponding anomaly scores for each snippet.

### 3.2 Multi-Timescale Feature Generator (MTFG)

Given $T$ snippets with a frame length of $L$ segmented from an input video, the MTFG generates three feature matrices of size $T \times D$, corresponding to the $D$-dimensional features of these $T$ snippets. Each feature matrix represents a temporal tubelet with the corresponding frame length. As illustrated in Fig.2a, a pre-trained Video Swin Transformer extracts features from a single snippet $t$ within a group of long temporal tubelets with a frame length $L_1$, medium temporal tubelets with a frame length $L_2$, and short temporal tubelets with a frame length $L_3$, resulting in $F_L'(t)$, $F_M'(t)$, and $F_S'(t)$ vectors, respectively. The vectors in each feature matrix are averaged along the temporal axis to form single feature vectors $F_L(t)$, $F_M(t)$, and $F_S(t)$. With $T$ snippets of the input video, the MTFG outputs three groups of $T$ feature vectors, i.e. $\{F_L(0), \cdots, F_L(T-1)\}$, and the other two with the same formulation $F_M$, and $F_S$ for medium and short, respectively. The $T$ feature vectors in each group are concatenated along the temporal axis, resulting in three feature matrices referring to $\boldsymbol{F_L}$, $\boldsymbol{F_M}$, and $\boldsymbol{F_S}$ in Fig.1 and Fig.2b.

### 3.3 Multi-Timescale Feature Fusion (MTFF)

The MTFF fuses the resulting feature maps according to their dependencies, as well as the local and global temporal correlation between snippets. As shown in Fig. 2b, the architecture consists of 4 fusion modules. Mod (i) Pairwise Feature Learning (PFL) employs three multi-head cross-attention (MCA) functions to perform pairwise fusion of 3 feature matrices $\boldsymbol{F_L}$, $\boldsymbol{F_M}$, and $\boldsymbol{F_S}$ generated by MTFG, and outputs $F_{LM}$, $F_{MS}$, and $F_{SL}$. Mod (ii) Local Temporal Correlation Learning (LTL) scales these pairwise fusion matrices by their local temporal correlation through 1D convolutional kernels. As depicted in mod (iii) Global Temporal Correlation Learning (GTL), $\boldsymbol{F_L}$, $\boldsymbol{F_M}$, and $\boldsymbol{F_S}$ are concatenated to form a feature matrix $\boldsymbol{F_{LMS_\text{concat}}}$. After the dimensional reduction of $F_{LMS_\text{concat}}$ by 1D convolution, we use the multi-head self-attention (MSA) block to scale the features based on the global temporal correlations across different snippets. Mod (iv) Feature Fusion (FF) concatenates the output feature maps of size $T \times D/2$ from mod's (ii) and (iii), and applies a 1D convolutional kernel to reduce the dimension. In the end, the MTFF outputs the final fused feature $\boldsymbol{X}$ as shown in Fig. 2b with a residual connection.

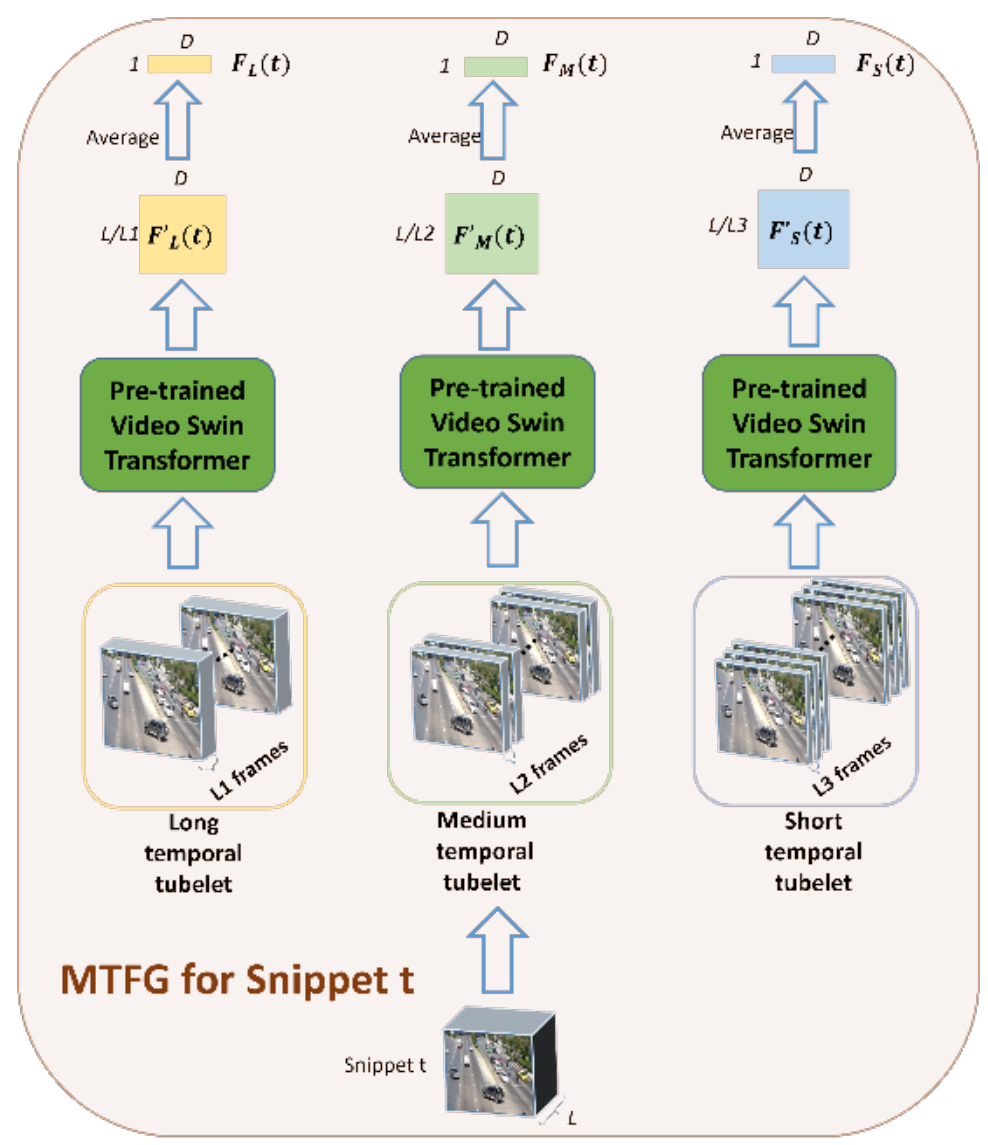

(a) Multi-Timescale Feature Generator.

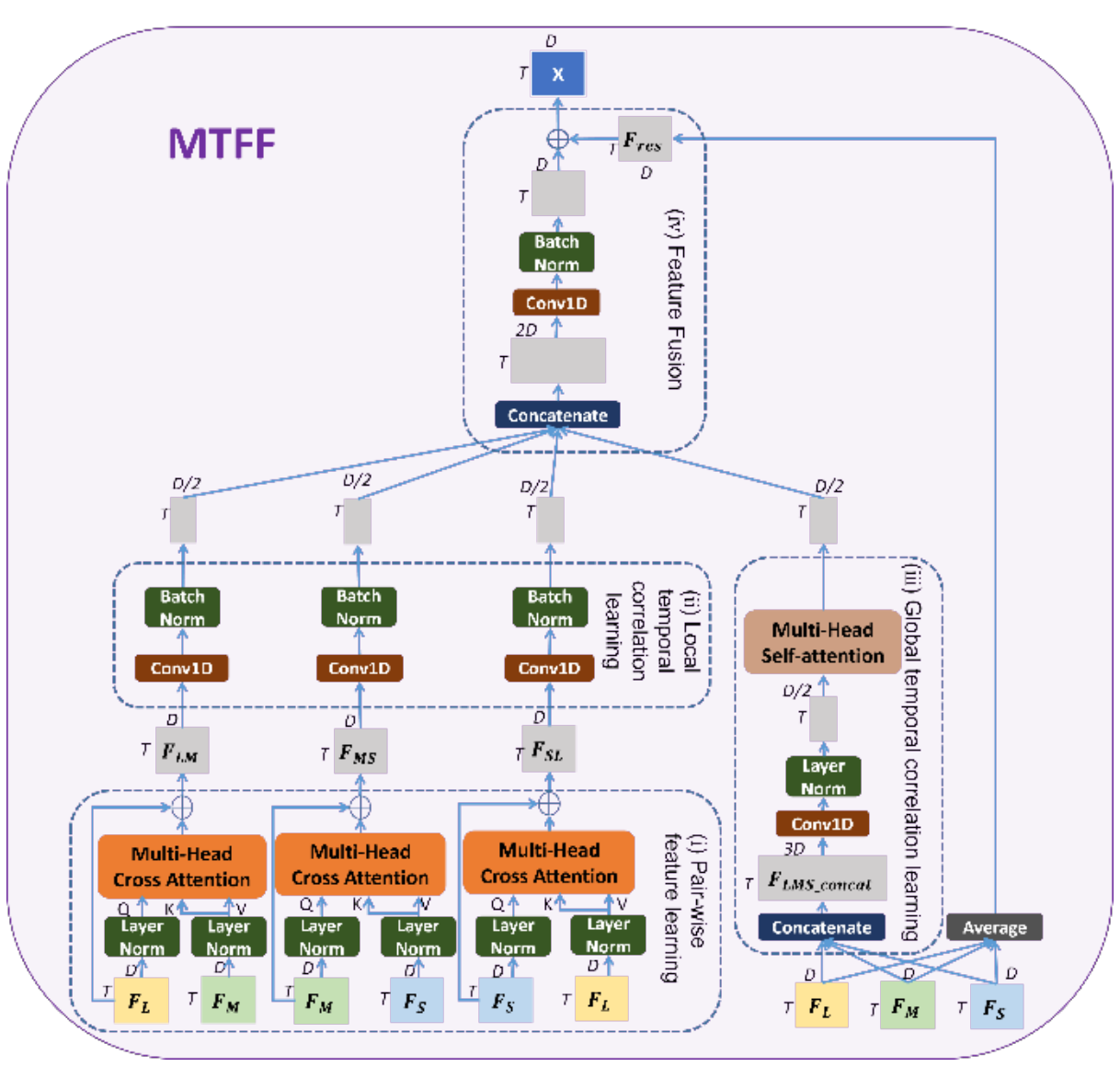

(b) Multi-Timescale Feature Fusion.

Figure 2. (a) MTFG (Multi-Timescale Feature Generator). Taking snippet t, a pre-trained Video Swin Transformer extracts features using 3 temporal tubelets with long, medium, and short frame lengths, i.e. $L_1$, $L_2$, and $L_3$, and the obtained features are averaged to 3 single feature vectors. (b) MTFF (Multi-Timescale Feature Fusion). Mod (i) learns the correlations between different path features for pairwise integration, mod (ii) captures local temporal dependencies and scales the features accordingly, and mod (iii) learns the global temporal dependencies of the integrated three features among $T$ snippets. Mod (iv) obtains the final feature representation **X** by fusing the output features from mod's (ii) and (iii).

Table 1: Number of samples for each class in the proposed VADD. The Video source column shows from which datasets the videos are taken. 1, 2, and 3 represent UCF-Crime [3], Throwing Action [4], and collected road-accident videos respectively. EMV refers to ”Enclosed Motor Vehicles” (cars and trunks), and VRU refers to ”Vulnerable Road Users” (motorbikes, bikes, and pedestrians).

| **Class name** | **# Training** | **# Testing** | **Video sources** |
|---|---|---|---|
| **Normal** | **900** | **170** | **1, 2, 3** |
| **Dangerous Throwing** | **154** | **25** | **2** |
| **Littering** | **81** | **12** | **2** |
| **VRU vs VRU** | **85** | **14** | **1, 3** |
| **EMV vs EMV** | **149** | **23** | **1, 3** |
| **EMV vs VRU** | **150** | **28** | **1, 3** |
| Abuse | 48 | 2 | 1 |
| Arrest | 45 | 5 | 1 |
| Arson | 41 | 9 | 1 |
| Assault | 47 | 3 | 1 |
| Burglary | 87 | 13 | 1 |
| Explosion | 29 | 21 | 1 |
| Fighting | 45 | 5 | 1 |
| Robbery | 145 | 5 | 1 |
| Shooting | 27 | 23 | 1 |
| Shoplifting | 29 | 21 | 1 |
| Stealing | 95 | 5 | 1 |
| Vandalism | 45 | 5 | 1 |

### 3.4 Anomaly Detection

After the MTFF, the fused feature matrix $\boldsymbol{X}$ is supplied into a 3-layer fully connected classifier to output the anomaly probability scores for $T$ snippets, as shown in Fig. 1. The loss function proposed by Tian et al. [8] is used for training the anomaly detection model, including loss functions for feature magnitude learning and classifier training, and a brief overview is provided in Eq. (1).

$$\mathcal{L}_{AD} = \mathcal{L}_{BCE} + \lambda_{fm}\mathcal{L}_{FM} + \lambda_1 \sum_{t=0}^{T-1} |s_t^+| + \lambda_2 \sum_{t=1}^{T-1} |s_t^+ - s_{t-1}^+|^2, \quad (1)$$

where $\mathcal{L}_{\mathcal{FM}}$ is the feature magnitude learning loss function [8] that maximizes the separability between the top-$k$ features from normal and abnormal videos, the binary cross-entropy (BCE) loss $\mathcal{L}_{BCE}$ is used for training the classifier, $\lambda_{fm}$, $\lambda_1$, and $\lambda_2$ are the weighting factors of each loss component, and $s_{t-1}^+$ and $s_t^+$ represent the scores of two consecutive snippets $t-1$ and $t$ in an anomalous input, respectively. The summations $\sum_{t=0}^{T-1} |s_t^+|$ and $\sum_{t=1}^{T-1} |s_t^+ - s_{t-1}^+|^2$ are added for sparsity and temporal smoothness constraints of scores in anomalous videos.

## 4. EXPERIMENTS

### 4.1 The Generated Dataset

The existing datasets for surveillance video anomaly detection are aiming at various scenarios and anomaly types. Some datasets contain only normal videos in their training sets and some abnormal events with annotations of bounding boxes in their testing sets. Other datasets like UCF-Crime [3], Throwing Action [4], and TAD [21], include various anomalous events with video-level labeling in both training and testing sets. For model development and covering more anomaly classes, we adopt the UCF-Crime dataset, which contains the most extensive range of abnormal event types and scenarios, and generate the Video Anomaly Detection Dataset (VADD) as an extended dataset. VADD includes 2,591 videos with a frame rate of 30 fps and a resolution of 320×240 pixels, with 2,202 train and 389 test videos. Compared to other datasets, the VADD provides the most general coverage of real-world anomalies, including one normal class, and 17 anomaly classes, as the details of video sources are given in Table 1. Train-set annotations only include a class label, while test-set annotations contain a video class label, a number of frames in a video, as well as the starting and ending frame positions of an abnormal event in a video.

### 4.2 Implementation Details

**MTFL method.** The MTFG module of the MTFL method, extracts video features from different frame-length tubelets. We apply frame lengths of $L_1 = 8$, $L_2 = 32$, $L_3 = 64$ (motivated by ablation studies) for short, medium, and long temporal tubelets, respectively. The sampling interval for all tubelets is 1 frame, and the output feature map of an input video comprises 32 feature vectors, corresponding to the total number of snippets ($T = 32$). For feature extraction, we utilize two pre-trained Video Swin Transformer models with different training methods: one pre-trained on the Kinetics-400 [22] and one pre-trained on the VADD with data augmentation involving horizontal flips and a 20% speed-up, namely VST-RGB and $\text{VST}_{\text{Aug}}$-RGB, respectively. In the MTFF module, we utilize three 3×1 Conv1D kernels for local temporal correlation learning, with mixed temporal reception achieved by various dilation rates (1, 2, and 4). The number of heads in the attention blocks is set to 4. The network is trained by the Adam optimizer with a weight decay of 0.0005 and a batch size of 128 for 1000 epochs. During training, each mini-batch consists of 64 randomly selected normal and 64 abnormal videos, with a learning rate of 0.0001.

**Evaluation Metrics.** Following the literature, the Receiver Operating Characteristic (ROC) curve and its corresponding Area Under the Curve (AUC) are employed for evaluation on the UCF-Crime, the ShanghaiTech, and the proposed VADD. Besides this, the Precision-Recall curve and its corresponding Average Precision (AP) are utilized to evaluate the performance of the proposed MTFL on the XD-Violence dataset [7].

## 4.3 Anomaly Detection Results

**Comparison to SotA methods.** Table 2 compares MTFL to SotA anomaly detection methods on the UCF-Crime, ShanghaiTech, and XD-Violence datasets. The MTFL method achieves 89.78% AUC on UCF-Crime, notably surpassing the SotA anomaly detection methods. The MTFL method demonstrates better capability for identifying complex anomalies with various real-world scenarios presented in the UCF-Crime and XD-Violence datasets, compared to simple anomalies with distinct motion patterns demonstrated in the ShanghaiTech dataset. Furthermore, in comparison to the methods utilizing the same VST-RGB features, such as MSL [9] and MGFN [13], MTFL exhibits a substantial superiority on the UCF-Crime and XD-Violence datasets, manifesting its efficiency regardless of the feature extraction approach. Moreover, contrasted with other weakly-supervised methods using the same loss function, the MTFL method outperforms RTFM [8] by 5.48% AUC on the UCF-Crime and 6.76% AP on the XD-Violence, and surpasses SSRL [10] by 2.35% AUC on the UCF-Crime. On the other hand, MTFL utilizing $\text{VST}_{\text{Aug}}$-RGB features achieves outstanding performance on the UCF-Crime, while MTFL using VST-RGB features shows more stable performance across all 3 datasets. This reveals that a $\text{VST}_{\text{Aug}}$-RGB pre-trained on anomaly video datasets can enhance feature representation of more complex real-world anomalies, while VST-RGB pre-trained on a general action dataset, such as Kinetics-400, demonstrates better generalization across different anomalous videos from any source, including movies or university campus scenes.

Table 2: Anomaly detection performance comparison with the SotA on UCF-Crime, ShanghaiTech, and XD-Violence.

| Supervision | Method | Year | Feature | UCF-Crime | ShanghaiTech | XD-Violence |
|---|---|---|---|---|---|---|
| Weakly | Sultani et al. [3] | 2018 | C3D-RGB | 75.41 | 86.30 | 73.20 |
| | GCN-Anomaly [23] | 2019 | TSN-RGB | 82.12 | 84.44 | - |
| | Wu et al. [7] | 2020 | I3D-RGB | 82.44 | - | 78.64 |
| | RTFM [8] | 2021 | I3D-RGB | 84.30 | 97.21 | 77.81 |
| | Wu et al. [24] | 2021 | I3D-RGB | 84.89 | 97.48 | 75.90 |
| | MSL [9] | 2022 | VST-RGB | 85.62 | 97.32 | 78.59 |
| | SSRL [10] | 2022 | I3D-RGB | 87.43 | **97.98** | - |
| | BN-WVAD [11] | 2023 | I3D-RGB | 87.24 | - | **84.93** |
| | MGFN [13] | 2023 | VST-RGB | 86.67 | - | 80.11 |
| | REWARD-E2E [18] | 2024 | Uniformer-32-RGB | 87.48 | - | 80.30 |
| | TPWNG [19] | 2024 | VIT-B/16 (RGB and Text) | 87.79 | - | 83.68 |
| Proposed | MTFL | 2024 | VST-RGB | 87.16 | 95.32 | 84.57 |
| | MTFL | 2024 | $\text{VST}_{\text{Aug}}$-RGB | **89.78** | 95.70 | 79.40 |

Table 3 shows the MTFL performance on the proposed VADD. We have retrained the RTFM [8] model using the same $\text{VST}_{\text{Aug}}$-RGB feature extraction, denoted by RTFM*. The proposed model achieves an AUC result of 88.42% on the VADD when detecting a wider range of anomalies, which is 0.62% higher than RTFM*.

Table 3: Detection performance on VADD.

| Method | Feature | VADD |
|---|---|---|
| RTFM* [8] | $\text{VST}_{\text{Aug}}$-RGB | 87.80 |
| **MTFL** | **$\text{VST}_{\text{Aug}}$-RGB** | **88.42** |

**Qualitative results**. Fig. 3 presents the anomaly scores generated by the MTFL model for randomly selected anomaly videos from the VADD, consisting of 1 normal and 7 anomalous videos. While 6 anomalous videos are sourced from the UCF-Crime, including *Arson016*, *Shooting004*, *Robbery102*, *Burglary037*, *Fighting033*, and *Shoplifting004*, one dangerous throwing video *CarDangerous024* is from the VADD. According to the anomaly scores, MTFL efficiently distinguishes between normal and anomalous segments in videos. It performs well even in scenarios where multiple events are occurring in a single video, such as *Burglary*. Apart from the detection of the anomalies with obvious motions, the MTFL demonstrates the ability to discern subtle anomalies that require an understanding of contextual information and behavior, like *Burglary* and *Shoplifting*. In the case of frequent throwing actions in *CarDangerous024*, the method cannot distinguish each individual throwing action, but it can still identify periods during which these actions occur.

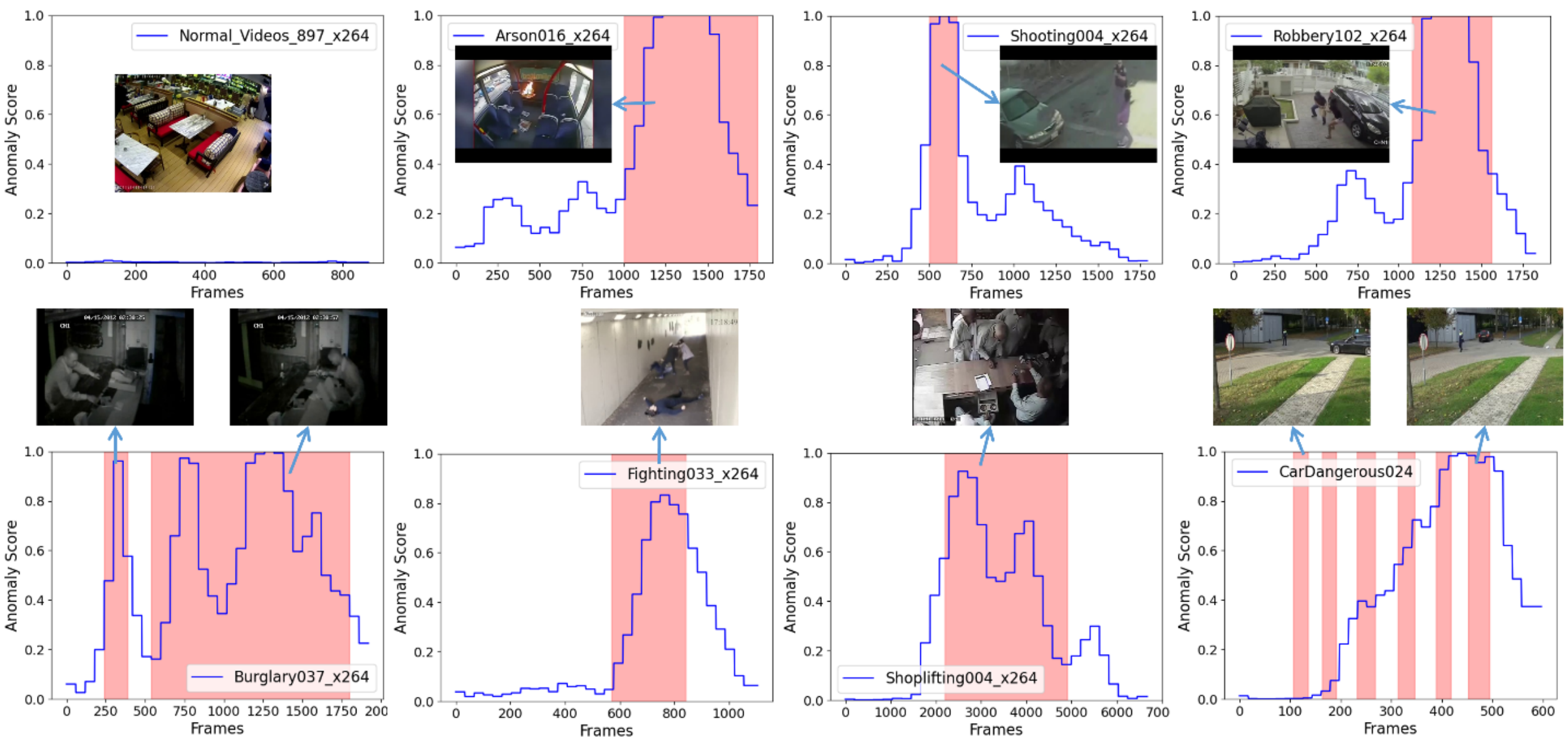

Figure 3. Anomaly scores of the proposed MTFL model on the UCF-Crime data (*Normal897*, *Arson016*, *Shooting004*, *Robbery102*, *Burglary037*, *Fighting033*, and *Shoplifting004*) and a dangerous throwing video from the VADD (*CarDangerous024*). The red-colored regions indicate the manually labeled occurrences of anomalous events.

## 5. ABLATION STUDIES

**Best feature extraction for anomaly detection.** To explore the impact of different feature extractors on the performance of anomaly detection, we examine five feature extraction models: C3D [5], I3D [6], SlowFast [25], MViTv2 [26], and VST [12] with RTFM method [8]. Table 4 shows the AUC results of anomaly detection using RTFM with these five video feature extractors on both UCF-Crime and VADD. Among the same anomaly detection methods, the VST model outperforms other feature extractors. When trained on the UCF-Crime dataset, the RTFM with VST results in 85.99% AUC on the UCF-Crime test set and 84.05% AUC on the VADD test set. With VADD-based training, applying the VST model within the RTFM yields 86.20% accuracy on the UCF-Crime test set and 85.75% accuracy on the VADD test set.

Table 4: Comparison of the RTFM method, employing five feature extractors trained on UCF-Crime and VADD.

| Data Source | Training Data | Architecture | UCF-Crime | VADD |
|---|---|---|---|---|
| Tian et al. [8] | UCF-Crime | I3D + RTFM | 84.30 | - |
| Our experiments | UCF-Crime | C3D + RTFM | 72.27 | 71.32 |
| | | I3D + RTFM | 82.63 | 81.58 |
| | | SlowFast + RTFM | 81.75 | 80.70 |
| | | MViTv2 + RTFM | 84.13 | 82.50 |
| | | **VST + RTFM** | **85.99** | **84.05** |
| | VADD | C3D + RTFM | 74.66 | 73.73 |
| | | I3D + RTFM | 83.75 | 82.97 |
| | | SlowFast + RTFM | 83.20 | 82.43 |
| | | MViTv2 + RTFM | 85.28 | 85.03 |
| | | VST + RTFM | 86.20 | 85.75 |
| | | **$\mathbf{VST_{Aug}}$ + RTFM** | **88.46** | **87.80** |

The experimental results indicate the benefits of VST features emphasizing local spatio-temporal correlations. Furthermore, we have trained the VST model on the proposed VADD with data augmentation via horizontal flips and speed up by 20%, resulting in the $VST_{Aug}$ model. Table 4 illustrates that the $VST_{Aug}$ model provides improvement by 2.26% on the UCF-Crime and 2.05% on the VADD, compared to the VST model results. Feature distribution reveals that the $VST_{Aug}$ model performs a more distinct clustering of video features. This distinctive feature distribution, combined with the sparsity-based anomaly detection process, enhances the separation between abnormal and normal segments better, compared to the VST model.

**Efficiency of internal MTFF stages.** Referring to Fig. 2b, the architecture of the designed MTFF module comprises four stages: PFL, LTL, GTL, and FF. To investigate the effects of these stages, we conduct experiments between different MTFF designs by removing different stages, as shown in Table 5. The results reveal that the most significant performance degradation is due to the removal of PFL and FF stages. This leads to noticeable drops in results on both UCF-Crime and VADD. The LTL and GTL stage removals result in relatively minor deterioration. These comparisons show that mutual learning of features extracted from spatio-temporal tubes of different frame lengths, coupled with feature fusion, enhances feature representation capability with the help of PFL and FF. Further enhancement in performance is achieved by incorporating local and global correlation learning to scale features within LTL and GTL. Overall, these experiments validate the impact of each stage in the MTFF module.

**Optimal combination of tubelets.** We extract features using 4 different tubelets with frame lengths of 8, 16, 32, and 64, respectively. By selecting three tubelet lengths as inputs to the MTFF module, we compare detection results for various length combinations as shown in Table 6. Notably, the best detection results are achieved when the short tubelet length is 8 frames, the medium tubelet length is 32 frames, and the long tubelet length is 64 frames. This observation suggests that combining the 8-frame feature for capturing rapid motion details, the 64-frame feature for contextual information, and the 32-frame feature for correlation information improves the understanding of anomalous events within the MTFL method.

Table 5: Ablation studies on four modules of MTFF.

| PFL | LTL | GTL | FF | UCF-Crime | VADD |
|---|---|---|---|---|---|
| | ✓ | ✓ | ✓ | 83.91 | 82.62 |
| ✓ | | ✓ | ✓ | 88.38 | 87.04 |
| ✓ | ✓ | | ✓ | 89.25 | 87.86 |
| ✓ | ✓ | ✓ | | 85.67 | 84.54 |
| ✓ | ✓ | ✓ | ✓ | **89.78** | **88.42** |

Table 6: Ablation studies on different tubelet lengths.

| 8 | 16 | 32 | 64 | UCF-Crime | VADD |
|---|---|---|---|---|---|
| | ✓ | ✓ | ✓ | 88.58 | 86.92 |
| ✓ | | ✓ | ✓ | **89.78** | **88.42** |
| ✓ | ✓ | | ✓ | 88.49 | 87.54 |
| ✓ | ✓ | ✓ | | 88.46 | 87.50 |

# 6. CONCLUSION

This paper has proposed an MTFL method that leverages multiple temporal scales to understand behaviour anomalies in videos, enabling a powerful fusion of motion details and event feature information for anomaly detection. The MTFL method achieves leading results on the UCF-Crime with 89.78% AUC and on the XD-Violence dataset with 84.57% AP, and presents comparable to SotA results on the ShanghaiTech dataset with 95.32% AUC. It shows high performance in detecting anomalies without distinct motion patterns due to the advanced time-scale partitioning. Moreover, to extend the available anomaly video data for development and experiments, we have generated VADD that enlarges the UCF-Crime dataset and covers a broader range of anomaly events. The MTFL method also showcases good results on the VADD.